\documentclass{article}

\usepackage{microtype}
\usepackage{subfiles}
\usepackage{graphicx}
\usepackage{subcaption}
\usepackage{lipsum}
\usepackage[table]{xcolor}
\usepackage{tabularx}
\usepackage{multirow}
\usepackage{booktabs}
\usepackage{soul}
\usepackage{xspace}
\usepackage{amsfonts}
\usepackage{amsmath}
\usepackage{algorithm}
\usepackage{algorithmic}
\usepackage{bbm}
\usepackage{mwe}
\usepackage{pifont}

\usepackage{hyperref}

\usepackage[accepted]{ool2020}

\newcolumntype{g}{>{\columncolor{gray!25}}c} 

\definecolor{lightblue}{HTML}{507CB8}
\newcommand{\primaltext}[1]{{#1}}

\newcommand{\cmark}{\ding{51}}%
\newcommand{\xmark}{\ding{55}}

\newcommand{\bcomment}[1]{}

\icmltitlerunning{Time for a Background Check! Uncovering the impact of Background Features on Deep Neural Networks}

\begin{document}

\twocolumn[
\icmltitle{Time for a Background Check! Uncovering the impact of Background Features on Deep Neural Networks}


\begin{icmlauthorlist}
\icmlauthor{Vikash Sehwag}{pri}
\icmlauthor{Rajvardhan Oak}{ber}
\icmlauthor{Mung Chiang}{pur}
\icmlauthor{Prateek Mittal}{pri}
\end{icmlauthorlist}

\icmlaffiliation{pri}{Princeton Universty, USA}
\icmlaffiliation{ber}{University of California, Berkeley, USA}
\icmlaffiliation{pur}{Purdue University, USA}

\icmlcorrespondingauthor{Vikash Sehwag}{vvikash@princeton.edu}

\vskip 0.3in
]

\printAffiliationsAndNotice{}

\vskip 0.3in

\begin{abstract}
  With increasing expressive power, deep neural networks have significantly improved the state-of-the-art on image classification datasets, such as ImageNet. In this paper, we investigate to what extent the increasing performance of deep neural networks is impacted by background features? In particular, we focus on \textit{background invariance}, i.e., accuracy unaffected by switching background features and \textit{background influence}, i.e., predictive power of background features itself when foreground is masked. We perform experiments with 32 different neural networks ranging from small-size networks~\cite{howard1905searching} to large-scale networks trained with up to one Billion images~\cite{yalniz2019billion}. Our investigations reveal that increasing expressive power of DNNs leads to higher influence of background features, while simultaneously, increases their ability to make the correct prediction when background features are removed or replaced with a randomly selected texture-based background.
\end{abstract}

\section{Introduction} \label{sec: introduction}
\begin{figure*}
    \centering
     \begin{subfigure}[b]{\textwidth}
         \centering
         \includegraphics[width=\linewidth]{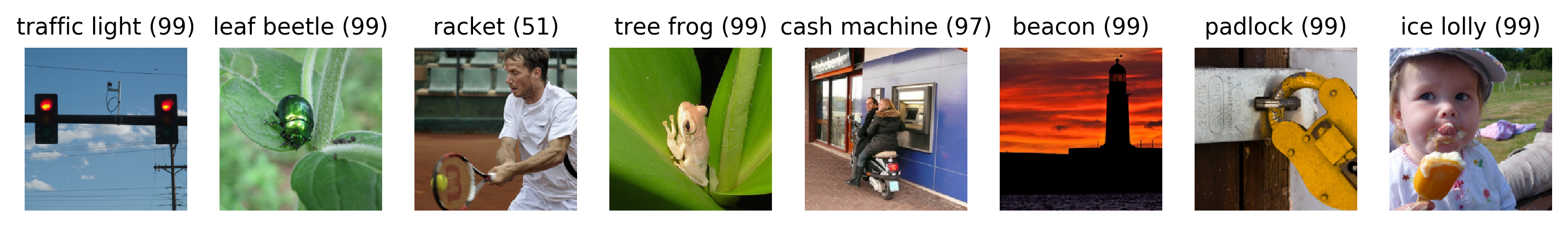}
         \caption{Output prediction on original images.}
     \end{subfigure}
     
    \begin{subfigure}[b]{\textwidth}
        \centering
        \includegraphics[width=\linewidth]{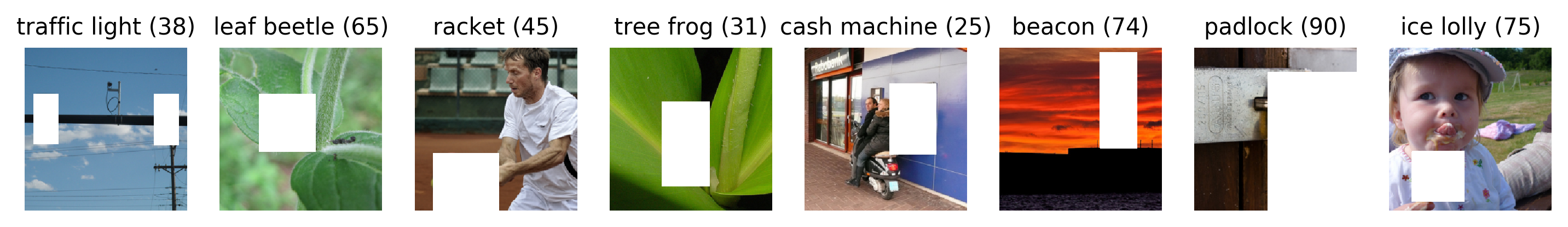}
        \caption{Prediction when foreground is whitened.}
    \end{subfigure}
     
    \caption{Predicted label of original and background only images (output confidence in parenthesis) for a ResNet-18 network trained on ImageNet dataset. It shows that even when the foreground is absent, the network is able to make a correct prediction for a large number of images. 
    Such correlation with background features can a sign of 1) a network utilizing context from background to make correct prediction or 2) a network exploiting the inherent bias in the data, thus using background correlation itself to discriminate between classes.}
    \label{fig: intro}
    \vspace{-10pt}
\end{figure*}

One key driver behind this success of modern deep neural networks (DNNs) is their expressive power, which enables them to learn a rich set of representations required to solve a target task~\cite{krizhevsky2012imagenet, bengio2013representation, he2016deep}. However, this expressive power raises a number of fundamental questions: how good are the learned representations? For example, does the learned representations correspond to semantic features in the image? Is the representation overfitted to the training data, or is it robust to the addition of semantically unrelated features in the image? 

In this paper, we argue that one fundamental principle for representation learning should be that the network should achieve background invariance~\cite{pinto2008real} i.e., be able to make a correct prediction when the background features are switched in an image classification task. Thus, we aim to understand to what extent current deep neural networks utilize the background information itself to solve the task of image classification. 

We develop a framework where we examine the influence of background image features on output prediction using two approaches: 1) We mask the foreground content of all test images, thus examine the correlation of output label with background features, 2) we switch the background of each image with an artificial background, such as white or texture-based images. In this setup, we expect the accuracy to remain unchanged assuming that the network has learned representations that are invariant to background changes. We rigorously test this framework on the ImageNet~\cite{deng2009imagenet} dataset where we evaluate the performance of 32 different DNNs proposed over the last couple of years~\cite{canziani2016analysis}. 

Our findings are intriguing: we observe that even on a diverse dataset like ImageNet, state-of-the-art DNNs ends up learning correlation with background signal. In fact, even when we mask the foreground, we find that DNNs can make a correct prediction for a significant number of images. We highlight a few such examples in Figure~\ref{fig: intro}, where just with the background content, a ResNet-18 network is able to make the correct prediction. We discuss this observation in more detail in Section~\ref{sec: exp-1}. Similarly, when we switch image backgrounds while keeping the foreground unchanged, we observe a significant reduction in the test accuracy. It further supports our hypothesis that current DNNs do really exploit background correlation to make the correct prediction. \primaltext{Although, with increasing expressive power, the drop in accuracy after a random background switch keeps decreasing.}  

Since the current training loss functions, such as cross-entropy loss~\cite{GoodBengCour16}, do not explicitly encourage background invariance, we find that increasing the expressive power of current networks further enables them to exploit any existing correlation between the background features and output prediction (supporting detail in Section~\ref{sec: exp-1}). Our results further question the ability of the deep neural networks to learn fundamental semantics, necessary to solve the task at hand, in the \textit{current training paradigms}. While further increasing the dataset size or diversity, thus training on a larger set of background variations, could be one approach to increase background invariance, we believe that a more successful approach will be to improve the training loss function to penalize correlation with the background. 

\noindent \textbf{Contributions.} We present a rigorous evaluation of 32 different DNNs to demonstrate that output prediction of deep neural networks is heavily influenced by background features. Our first set of results shows that with increasing expressive power, the tendency of DNNs to exploit background features to solve the task at hand increases. Next, we demonstrate a large reduction in the performance of DNNS when we switch the background features across multiple datasets and network architectures. \primaltext{However, this gap in performance decreases with increasing expressive power of DNNs.} 

\section{Methods and experimental setup} \label{sec: method}
Most computer vision datasets, such as ImageNet, VOC12, Caltech101 have object categories corresponding to a noun. This structure enables us to divide each image into two parts: Foreground, which comprises the correct objects, and background, which is everything in the image except the foreground. Note that for datasets based on scenes (SUN~\cite{xiao2010sun}), texture (DTD~\cite{cimpoi14dtd}) such categorization of foreground and background is not feasible. 

Our objective is to analyze the following question: To what extent current deep neural networks exploit background features to achieve the targeted classification. We approach this question from two directions.

\noindent \textbf{Testing background influence by masking foreground.} We first aim to explicitly measure the correlation between background features and output prediction. We mask the foreground, i.e., the object corresponding to correct prediction and replace it with a white patch. We choose a white patch, instead of random noise to have a minimal influence of the patch on output. We refer to such images as background images. 

\noindent \textbf{Testing background invariance by switching background.} Now we aim to analyze the effect of the absence of background features. We argue that if the learned representations are agnostic to the background features, replacing it with a white image or other novel patterns like texture, outdoor scenes won't affect it. Thus in this setup, we expect networks to achieve accuracy close to unmodified images. 

We also focus on relative accuracy, instead of an absolute number for accuracy, for both of the aforementioned tasks. We argue that it potentially allows us the analyze the performance on images unaffected by labeling noise in the foreground and background classification. We discuss this choice in detail in Appendix~\ref{app: baseline}.

\begin{table*}
    \vspace{-10pt}
    \caption{Analyzing the correlation between the expressive power of a neural network and its tendency to latch on background information to make correct prediction. \cmark and \xmark~implies whether the background features are itself classified as the true class or not. It shows that with increasing expressive power, we found a new set of images, where the background features are sufficient to achieve correct classification for networks with equal or higher expressive power, but not for networks with lower expressive power.}
    \label{tab: vary_exppower}
    \centering
    \resizebox{0.95\textwidth}{!}{
    \begin{tabular}{cccp{5pt}p{30pt}p{30pt}p{30pt}p{30pt}p{30pt}p{30pt}p{30pt}}
        \toprule
        Architecture & \begin{tabular}[c]{@{}c@{}}Accuracy on \\ unmodified images (\%)\end{tabular} & \begin{tabular}[c]{@{}c@{}}Accuracy with \\ whitened foreground (\%)\end{tabular}& \multicolumn{8}{c}{\includegraphics[width=0.58\linewidth]{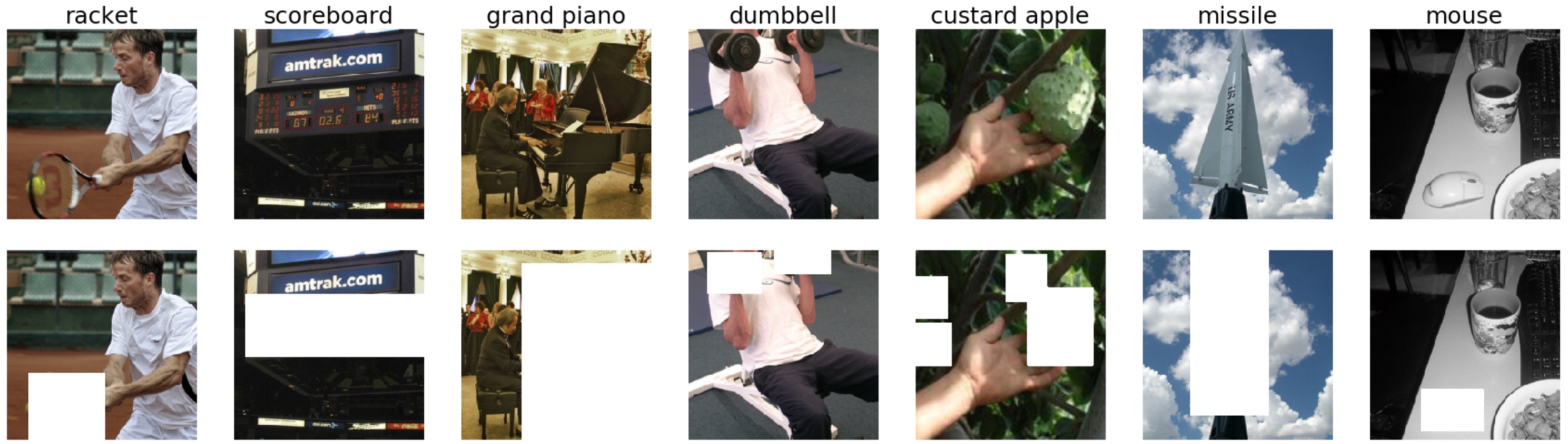}} \\ \midrule
        ResNet-18~\cite{he2016deep} & 74.3 & 0.89 & & \cmark & \xmark & \xmark & \xmark & \xmark & \xmark & \xmark \\
        MobileNet-v3~\cite{howard1905searching} & 76.5 & 0.93 & & \cmark & \cmark & \xmark & \xmark & \xmark & \xmark & \xmark \\
        ResNet101~\cite{he2016deep} & 80.7 & 1.55 & & \cmark & \cmark & \cmark & \xmark & \xmark & \xmark & \xmark \\
        Wide-ResNet-50-2~\cite{zagoruyko2016wide} & 81.7 & 2.28 & & \cmark & \cmark & \cmark & \cmark & \xmark & \xmark & \xmark \\
        EfficientNet-b1~\cite{tan2019efficientnet} & 83.6 & 3.78 & & \cmark & \cmark & \cmark & \cmark & \cmark & \xmark & \xmark  \\
        Swsl-ResNext50-32x4d~\cite{yalniz2019billion} & 84.6 & 3.75 & & \cmark & \cmark & \cmark & \cmark & \cmark & \cmark & \xmark  \\
        ResNext101-32x48d~\cite{mahajan2018exploring} & 87.4 & 5.88 & & \cmark & \cmark & \cmark & \cmark & \cmark & \cmark & \cmark \\ \bottomrule
    \end{tabular}
    }
    \vspace{-10pt}
\end{table*}

\noindent \textbf{Experimental setup.} We use 31,801 images from the validation set of ImageNet dataset (data preparation process described in Appendix~\ref{app: dataprep}). We evaluate 32 different networks\footnote{we use publicly available checkpoints from \url{https://github.com/rwightman/pytorch-image-models}.} in our framework. These networks bring in variation in depth (ResNet-18 to resnet101), variation in architecture (ResNet, DPN, MobileNet, etc)~\cite{he2016deep, chen2017dual, howard1905searching}), and effect of training setup (models trained on $\sim$ 1 Billion images using the semi-weakly supervised training~\cite{yalniz2019billion, mahajan2018exploring}). We primarily use white and texture-based images as two alternatives to replace the image background. We demonstrate a similar trend with other choices such as background based on Gaussian or uniform noise or any random color. Similarly, we also show a similar trend as ImageNet for Caltech101 and VOC12 dataset. We use the ResNet18 network as the baseline network.

\section{Background Influence: DNNs learns correlation between correct labels and background features} \label{sec: exp-1}
\begin{figure}
    \centering
    \includegraphics[width=\linewidth]{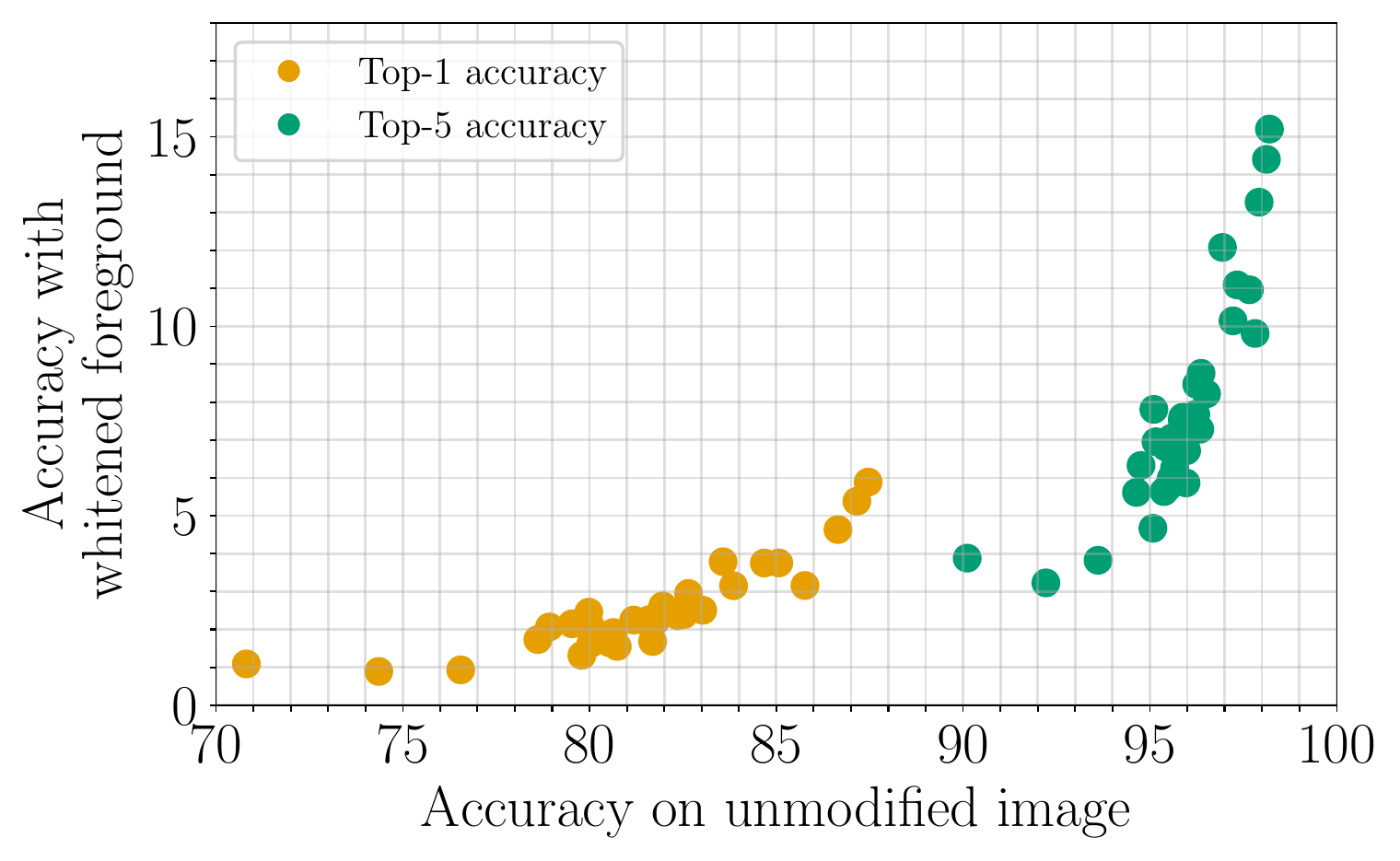}
    \caption{Top-1 and top-5 accuracy for background features when the foreground content is masked. Each dot corresponds to one neural network. It shows that with increasing expressive power, current DNNs increasingly exploit correlation of background features and correct prediction.}
    \label{fig: back_only}
    \vspace{-10pt}
\end{figure}

Now we present the results with foreground masking, where we mask the foreground and test whether deep neural networks have learned correlation with background features. 

Figure~\ref{fig: back_only} presents these results where we measure both top-1 and top-5 accuracy of 32 different networks on only background features. We plot these results in order of their test accuracy on unmodified ImageNet images and observe that the influence of background features increases with the expressive power of DNNs. Such correlation with background features can a sign of 1) either a network utilizing context from background to make a correct prediction or 2) or a network exploiting the inherent bias in the data, thus using background correlation itself to discriminate between classes.

We analyze this phenomenon in more detail in Table~\ref{tab: vary_exppower}. We consider seven networks with different expressive power and report a set of images where they make correction predictions solely based on background features. It shows that there is a common set of images where all networks are able to make a correct prediction with background features. However, as we increase expressive power, we find a novel subset of images for which background features are not predictive for any less expressive networks but able to make the right prediction for all more expressive networks. 

\section{Background invariance: Switching background leads to performance degradation} \label{sec: exp-2}
\begin{figure*}
    \centering
    \begin{subfigure}[b]{0.48\textwidth}
        \centering
        \includegraphics[width=\textwidth]{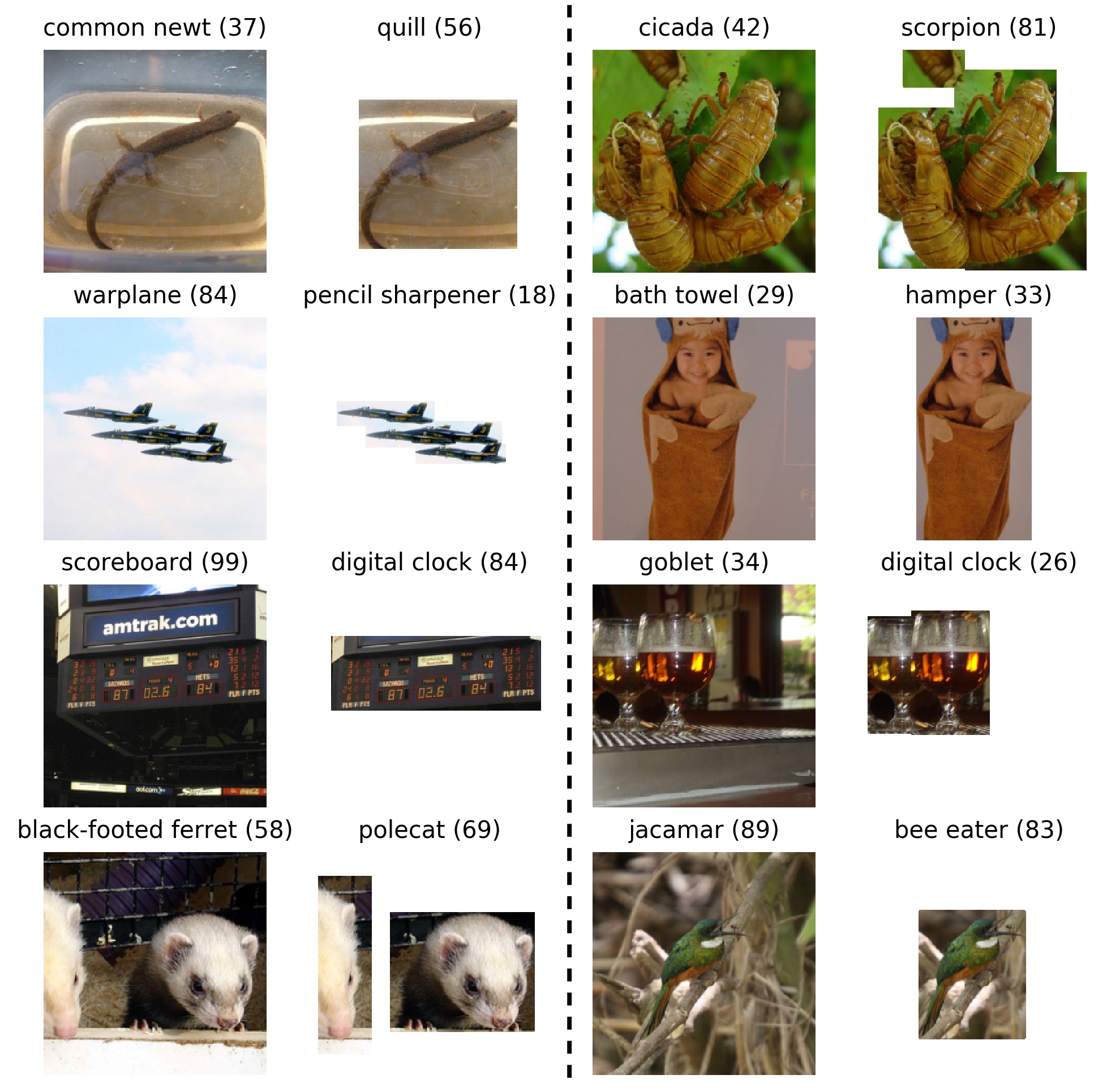}
        \caption{Using white background}
        \label{fig: demo_backswitch_white}
    \end{subfigure}
    \hfill
    \begin{subfigure}[b]{0.48\textwidth}
        \centering
        \includegraphics[width=\textwidth]{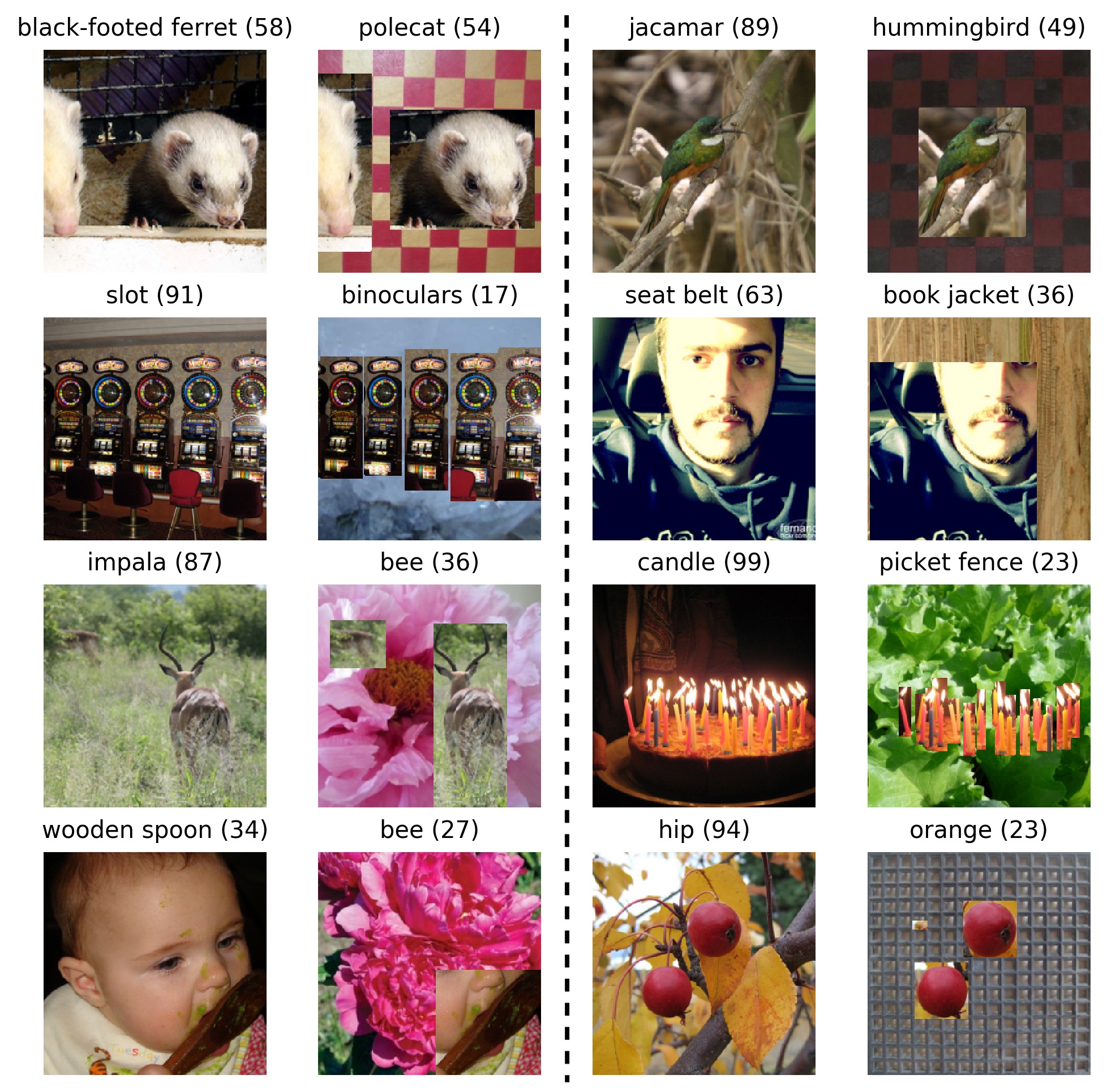}
        \caption{Using texture-based background}
        \label{fig: demo_backswitch_dtd}
    \end{subfigure}
    \vspace{-5pt}
    \caption{Mis-classification for ResNet-18 network when we change the background features.}
    \label{fig: demo_backswitch}
    \vspace{-10pt}
\end{figure*}

Here we discuss the results with the background switching experiment where we replace the original background features with white or texture-based images. 

\begin{figure}[!htb]
    \centering
    \includegraphics[width=\linewidth]{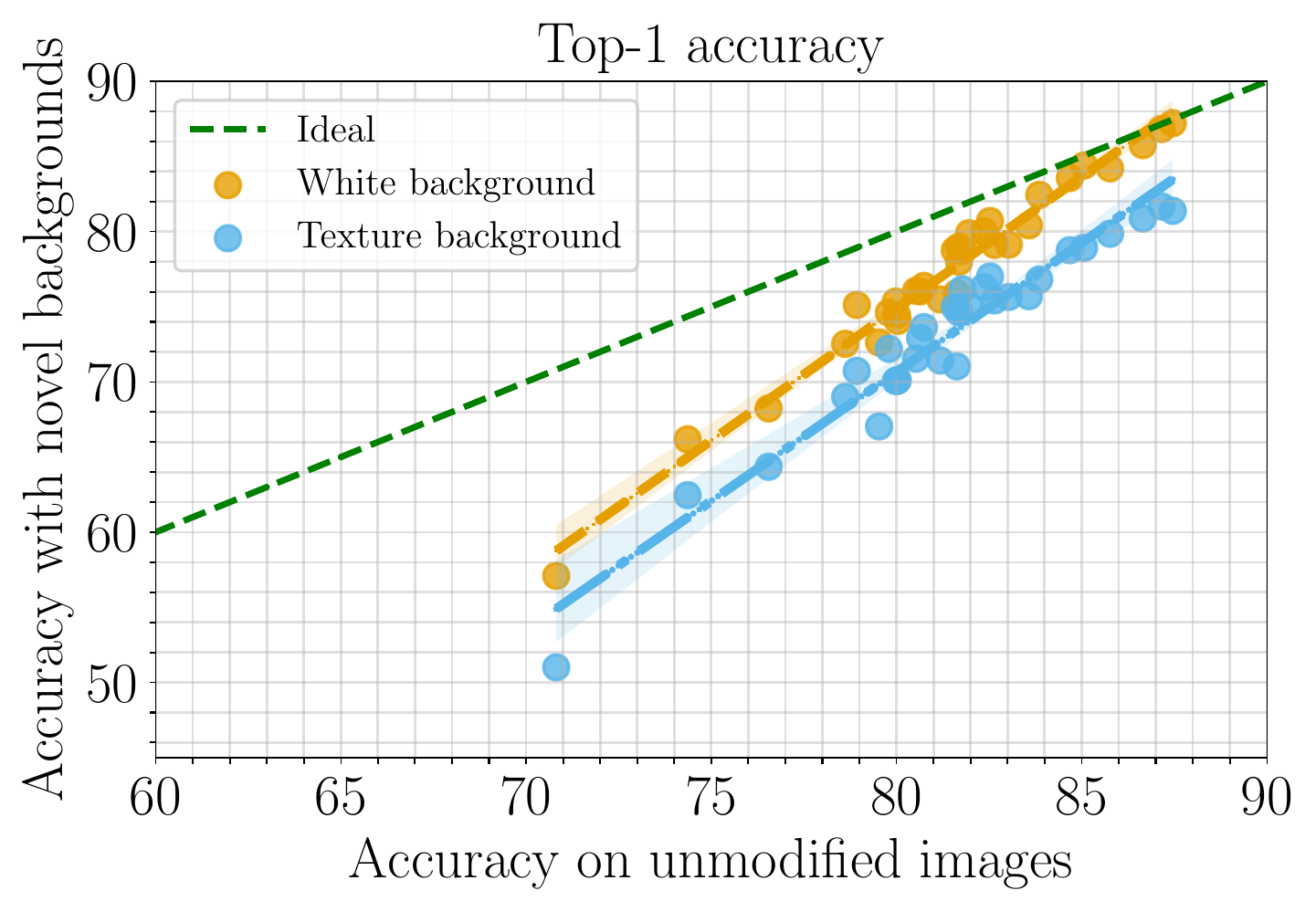}
    \caption{Top-1 accuracy on ImageNet test images when the background is switched to white or texture based background. Each dot correspond to one neural network.}
    \label{fig: switch_back_main}
\end{figure}

Figure~\ref{fig: switch_back_main} presents the top-1 accuracy for all 32 networks evaluated in this experiment (top-5 accuracy in Appendix~\ref{app: background_switch_top5}). If the networks had learned a set of good representation, the accuracy should not change with a change in the background. However, our results show a large drop in performance when we switch background features. However, with increase expressive power of DNNs, we observe that this gap decreases. We present a few examples demonstrating this phenomenon in Figure~\ref{fig: demo_backswitch}. 

When tested with the baseline ResNet-18 network, we observe a similar drop in performance for Gaussian noise, Uniform noise, green, yellow, grey color-based backgrounds. Similarly, for the VOC12 dataset~\cite{everingham2015voc12}, test accuracy decreases to 75\% and 46\% when using white and texture-based backgrounds, respectively. Original images achieved 95\% test accuracy on the same dataset. For the Caltech101 dataset~\cite{fei2004caltech101, fei2006caltech101}, test accuracy decrease to 85\% and 74\%, when using white and texture-based backgrounds, respectively. We report the mean accuracy over 3 runs. Original images achieved 94\% test accuracy on the same dataset. 

\section{Related work} \label{sec: rel_work}
Our work is closely related to an existing line of research~\cite{zhu2016WorWoObject, torralba2003contextualPriming, sun2017SeeingNotThere, katti2019humancontextbias, geirhos2020shortcutLearning}, and a concurrent work~\cite{xiao2020NoiseOrSingal}, aiming to understand the relationship of foreground and background features with network performance. However, our work particularly delves deeper into understanding this relationship with the increasing expressive power of DNNs. Another related line of research in biological vision studies the importance of context~\cite{oliva2007role, hock1974contextual, biederman1982scene, hayes2007effect} and other studies the impact of changing backgrounds~\cite{dicarlo2012does, yamins2014performance, pinto2008real, george2009towards}.

\section{Discussion}
In this work, we demonstrate a significant impact of background features on the performance of deep neural networks. We provide supporting experimental results in two frameworks:  Evaluating performance 1) only on background features 2) while removing or switching background features. With increasing expressive power of DNNs, we observe an increasing tendency to exploit background features to get the correct prediction, while simultaneously increasing the ability to make a correct prediction with foreground features only. Future works should aim to delve deeper into the correlation with background features while also measuring the performance with the worst-case background, instead of a random background.

\bibliography{ref}

\begin{thebibliography}{31}
\providecommand{\natexlab}[1]{#1}
\providecommand{\url}[1]{\texttt{#1}}
\expandafter\ifx\csname urlstyle\endcsname\relax
  \providecommand{\doi}[1]{doi: #1}\else
  \providecommand{\doi}{doi: \begingroup \urlstyle{rm}\Url}\fi

\bibitem[Bengio et~al.(2013)Bengio, Courville, and
  Vincent]{bengio2013representation}
Bengio, Y., Courville, A., and Vincent, P.
\newblock Representation learning: A review and new perspectives.
\newblock \emph{IEEE transactions on pattern analysis and machine
  intelligence}, 35\penalty0 (8):\penalty0 1798--1828, 2013.

\bibitem[Biederman et~al.(1982)Biederman, Mezzanotte, and
  Rabinowitz]{biederman1982scene}
Biederman, I., Mezzanotte, R.~J., and Rabinowitz, J.~C.
\newblock Scene perception: Detecting and judging objects undergoing relational
  violations.
\newblock \emph{Cognitive psychology}, 14\penalty0 (2):\penalty0 143--177,
  1982.

\bibitem[Canziani et~al.(2016)Canziani, Paszke, and
  Culurciello]{canziani2016analysis}
Canziani, A., Paszke, A., and Culurciello, E.
\newblock An analysis of deep neural network models for practical applications.
\newblock \emph{arXiv preprint arXiv:1605.07678}, 2016.

\bibitem[Chen et~al.(2017)Chen, Li, Xiao, Jin, Yan, and Feng]{chen2017dual}
Chen, Y., Li, J., Xiao, H., Jin, X., Yan, S., and Feng, J.
\newblock Dual path networks.
\newblock In \emph{Advances in neural information processing systems}, pp.\
  4467--4475, 2017.

\bibitem[Cimpoi et~al.(2014)Cimpoi, Maji, Kokkinos, Mohamed, , and
  Vedaldi]{cimpoi14dtd}
Cimpoi, M., Maji, S., Kokkinos, I., Mohamed, S., , and Vedaldi, A.
\newblock Describing textures in the wild.
\newblock In \emph{Proceedings of the {IEEE} Conf. on Computer Vision and
  Pattern Recognition ({CVPR})}, 2014.

\bibitem[Deng et~al.(2009)Deng, Dong, Socher, Li, Li, and
  Fei-Fei]{deng2009imagenet}
Deng, J., Dong, W., Socher, R., Li, L.-J., Li, K., and Fei-Fei, L.
\newblock Imagenet: A large-scale hierarchical image database.
\newblock In \emph{2009 IEEE conference on computer vision and pattern
  recognition}, pp.\  248--255. Ieee, 2009.

\bibitem[DiCarlo et~al.(2012)DiCarlo, Zoccolan, and Rust]{dicarlo2012does}
DiCarlo, J.~J., Zoccolan, D., and Rust, N.~C.
\newblock How does the brain solve visual object recognition?
\newblock \emph{Neuron}, 73\penalty0 (3):\penalty0 415--434, 2012.

\bibitem[Everingham et~al.(2015)Everingham, Eslami, Van~Gool, Williams, Winn,
  and Zisserman]{everingham2015voc12}
Everingham, M., Eslami, S.~A., Van~Gool, L., Williams, C.~K., Winn, J., and
  Zisserman, A.
\newblock The pascal visual object classes challenge: A retrospective.
\newblock \emph{International journal of computer vision}, 111\penalty0
  (1):\penalty0 98--136, 2015.

\bibitem[Fei-Fei et~al.(2004)Fei-Fei, Fergus, and Perona]{fei2004caltech101}
Fei-Fei, L., Fergus, R., and Perona, P.
\newblock Learning generative visual models from few training examples: An
  incremental bayesian approach tested on 101 object categories.
\newblock In \emph{2004 conference on computer vision and pattern recognition
  workshop}, pp.\  178--178. IEEE, 2004.

\bibitem[Fei-Fei et~al.(2006)Fei-Fei, Fergus, and Perona]{fei2006caltech101}
Fei-Fei, L., Fergus, R., and Perona, P.
\newblock One-shot learning of object categories.
\newblock \emph{IEEE transactions on pattern analysis and machine
  intelligence}, 28\penalty0 (4):\penalty0 594--611, 2006.

\bibitem[Geirhos et~al.(2020)Geirhos, Jacobsen, Michaelis, Zemel, Brendel,
  Bethge, and Wichmann]{geirhos2020shortcutLearning}
Geirhos, R., Jacobsen, J.-H., Michaelis, C., Zemel, R., Brendel, W., Bethge,
  M., and Wichmann, F.~A.
\newblock Shortcut learning in deep neural networks.
\newblock \emph{arXiv preprint arXiv:2004.07780}, 2020.

\bibitem[George \& Hawkins(2009)George and Hawkins]{george2009towards}
George, D. and Hawkins, J.
\newblock Towards a mathematical theory of cortical micro-circuits.
\newblock \emph{PLoS computational biology}, 5\penalty0 (10), 2009.

\bibitem[Goodfellow et~al.(2016)Goodfellow, Bengio, and
  Courville]{GoodBengCour16}
Goodfellow, I.~J., Bengio, Y., and Courville, A.
\newblock \emph{Deep Learning}.
\newblock MIT Press, Cambridge, MA, USA, 2016.
\newblock \url{http://www.deeplearningbook.org}.

\bibitem[Hayes et~al.(2007)Hayes, Nadel, and Ryan]{hayes2007effect}
Hayes, S.~M., Nadel, L., and Ryan, L.
\newblock The effect of scene context on episodic object recognition:
  parahippocampal cortex mediates memory encoding and retrieval success.
\newblock \emph{Hippocampus}, 17\penalty0 (9):\penalty0 873--889, 2007.

\bibitem[He et~al.(2016)He, Zhang, Ren, and Sun]{he2016deep}
He, K., Zhang, X., Ren, S., and Sun, J.
\newblock Deep residual learning for image recognition.
\newblock In \emph{Proceedings of the IEEE conference on computer vision and
  pattern recognition}, pp.\  770--778, 2016.

\bibitem[Hock et~al.(1974)Hock, Gordon, and Whitehurst]{hock1974contextual}
Hock, H.~S., Gordon, G.~P., and Whitehurst, R.
\newblock Contextual relations: the influence of familiarity, physical
  plausibility, and belongingness.
\newblock \emph{Perception \& Psychophysics}, 16\penalty0 (1):\penalty0 4--8,
  1974.

\bibitem[Howard et~al.(2019)Howard, Sandler, Chu, Chen, Chen, Tan, Wang, Zhu,
  Pang, Vasudevan, et~al.]{howard1905searching}
Howard, A., Sandler, M., Chu, G., Chen, L., Chen, B., Tan, M., Wang, W., Zhu,
  Y., Pang, R., Vasudevan, V., et~al.
\newblock Searching for mobilenetv3. arxiv 2019.
\newblock \emph{arXiv preprint arXiv:1905.02244}, 2019.

\bibitem[Katti et~al.(2019)Katti, Peelen, and Arun]{katti2019humancontextbias}
Katti, H., Peelen, M.~V., and Arun, S.
\newblock Machine vision benefits from human contextual expectations.
\newblock \emph{Scientific reports}, 9\penalty0 (1):\penalty0 1--12, 2019.

\bibitem[Krizhevsky et~al.(2012)Krizhevsky, Sutskever, and
  Hinton]{krizhevsky2012imagenet}
Krizhevsky, A., Sutskever, I., and Hinton, G.~E.
\newblock Imagenet classification with deep convolutional neural networks.
\newblock In \emph{Advances in neural information processing systems}, pp.\
  1097--1105, 2012.

\bibitem[Mahajan et~al.(2018)Mahajan, Girshick, Ramanathan, He, Paluri, Li,
  Bharambe, and van~der Maaten]{mahajan2018exploring}
Mahajan, D., Girshick, R., Ramanathan, V., He, K., Paluri, M., Li, Y.,
  Bharambe, A., and van~der Maaten, L.
\newblock Exploring the limits of weakly supervised pretraining.
\newblock In \emph{Proceedings of the European Conference on Computer Vision
  (ECCV)}, pp.\  181--196, 2018.

\bibitem[Oliva \& Torralba(2007)Oliva and Torralba]{oliva2007role}
Oliva, A. and Torralba, A.
\newblock The role of context in object recognition.
\newblock \emph{Trends in cognitive sciences}, 11\penalty0 (12):\penalty0
  520--527, 2007.

\bibitem[Pinto et~al.(2008)Pinto, Cox, and DiCarlo]{pinto2008real}
Pinto, N., Cox, D.~D., and DiCarlo, J.~J.
\newblock Why is real-world visual object recognition hard?
\newblock \emph{PLoS computational biology}, 4\penalty0 (1), 2008.

\bibitem[Sun \& Jacobs(2017)Sun and Jacobs]{sun2017SeeingNotThere}
Sun, J. and Jacobs, D.~W.
\newblock Seeing what is not there: Learning context to determine where objects
  are missing.
\newblock In \emph{Proceedings of the IEEE Conference on Computer Vision and
  Pattern Recognition}, pp.\  5716--5724, 2017.

\bibitem[Tan \& Le(2019)Tan and Le]{tan2019efficientnet}
Tan, M. and Le, Q.~V.
\newblock Efficientnet: Rethinking model scaling for convolutional neural
  networks.
\newblock \emph{arXiv preprint arXiv:1905.11946}, 2019.

\bibitem[Torralba(2003)]{torralba2003contextualPriming}
Torralba, A.
\newblock Contextual priming for object detection.
\newblock \emph{International journal of computer vision}, 53\penalty0
  (2):\penalty0 169--191, 2003.

\bibitem[Xiao et~al.(2010)Xiao, Hays, Ehinger, Oliva, and
  Torralba]{xiao2010sun}
Xiao, J., Hays, J., Ehinger, K.~A., Oliva, A., and Torralba, A.
\newblock Sun database: Large-scale scene recognition from abbey to zoo.
\newblock In \emph{2010 IEEE Computer Society Conference on Computer Vision and
  Pattern Recognition}, pp.\  3485--3492. IEEE, 2010.

\bibitem[Xiao et~al.(2020)Xiao, Engstrom, Ilyas, and
  Madry]{xiao2020NoiseOrSingal}
Xiao, K., Engstrom, L., Ilyas, A., and Madry, A.
\newblock Noise or signal: The role of image backgrounds in object recognition.
\newblock \emph{arXiv preprint arXiv:2006.09994}, 2020.

\bibitem[Yalniz et~al.(2019)Yalniz, J{\'e}gou, Chen, Paluri, and
  Mahajan]{yalniz2019billion}
Yalniz, I.~Z., J{\'e}gou, H., Chen, K., Paluri, M., and Mahajan, D.
\newblock Billion-scale semi-supervised learning for image classification.
\newblock \emph{arXiv preprint arXiv:1905.00546}, 2019.

\bibitem[Yamins et~al.(2014)Yamins, Hong, Cadieu, Solomon, Seibert, and
  DiCarlo]{yamins2014performance}
Yamins, D.~L., Hong, H., Cadieu, C.~F., Solomon, E.~A., Seibert, D., and
  DiCarlo, J.~J.
\newblock Performance-optimized hierarchical models predict neural responses in
  higher visual cortex.
\newblock \emph{Proceedings of the National Academy of Sciences}, 111\penalty0
  (23):\penalty0 8619--8624, 2014.

\bibitem[Zagoruyko \& Komodakis(2016)Zagoruyko and
  Komodakis]{zagoruyko2016wide}
Zagoruyko, S. and Komodakis, N.
\newblock Wide residual networks.
\newblock \emph{arXiv preprint arXiv:1605.07146}, 2016.

\bibitem[Zhu et~al.(2016)Zhu, Xie, and Yuille]{zhu2016WorWoObject}
Zhu, Z., Xie, L., and Yuille, A.~L.
\newblock Object recognition with and without objects.
\newblock \emph{arXiv preprint arXiv:1611.06596}, 2016.

\end{thebibliography}
\bibliographystyle{ool2020}

\appendix

\section{Setting up a baseline} \label{app: baseline}
ImageNet, the dataset we use in this work, does include some label noise in bounding boxes for background and foreground classification. For some images, we find that not all foreground content is labeled as foreground, e.g., not all ants in an image are labeled as foreground, thus even after removing foreground, the background image still has traces of the correct object. In that case, it is expected to obtain correct classification even with the background image. Similarly, if a significant part of the foreground image is dropped due to poor labeling, the network might not be able to achieve the correct classification with a foreground image. 

To address this challenge, we focus on relative performance in our framework w.r.t. to a baseline network. We assume that any error on bounding boxes for foreground objects will affect the performance of all networks. Thus relative performance w.r.t. to a baseline network, potentially allows us the analyze the performance on images unaffected by labeling noise. 

\section{Data Preparation} \label{app: dataprep}
We use the images from the validation set of ImageNet dataset (50,000 images) following a two-step filtering process. First, we remove images with foreground area less than 5\% of the image (for classes like ping-pong ball). With such images, when we mask the foreground, most networks still make the correct prediction. Similarly, with only foreground, most networks fail to make a correct prediction. 

Next, we remove images where the foreground area is more than 70\% of the image. We aim to understand the impact of background features where it desirable to have a significant ratio of background features in the image. Otherwise, if the image foreground covers the whole image, it is unlikely that background switching will cause misclassification. 

Following this filtering process, we arrive at 31,801 images which we use throughout our experiments in this paper. 

\section{Additional results for background switching experiment} \label{app: background_switch_top5}

\begin{figure}[H]
    \centering
    \includegraphics[width=\linewidth]{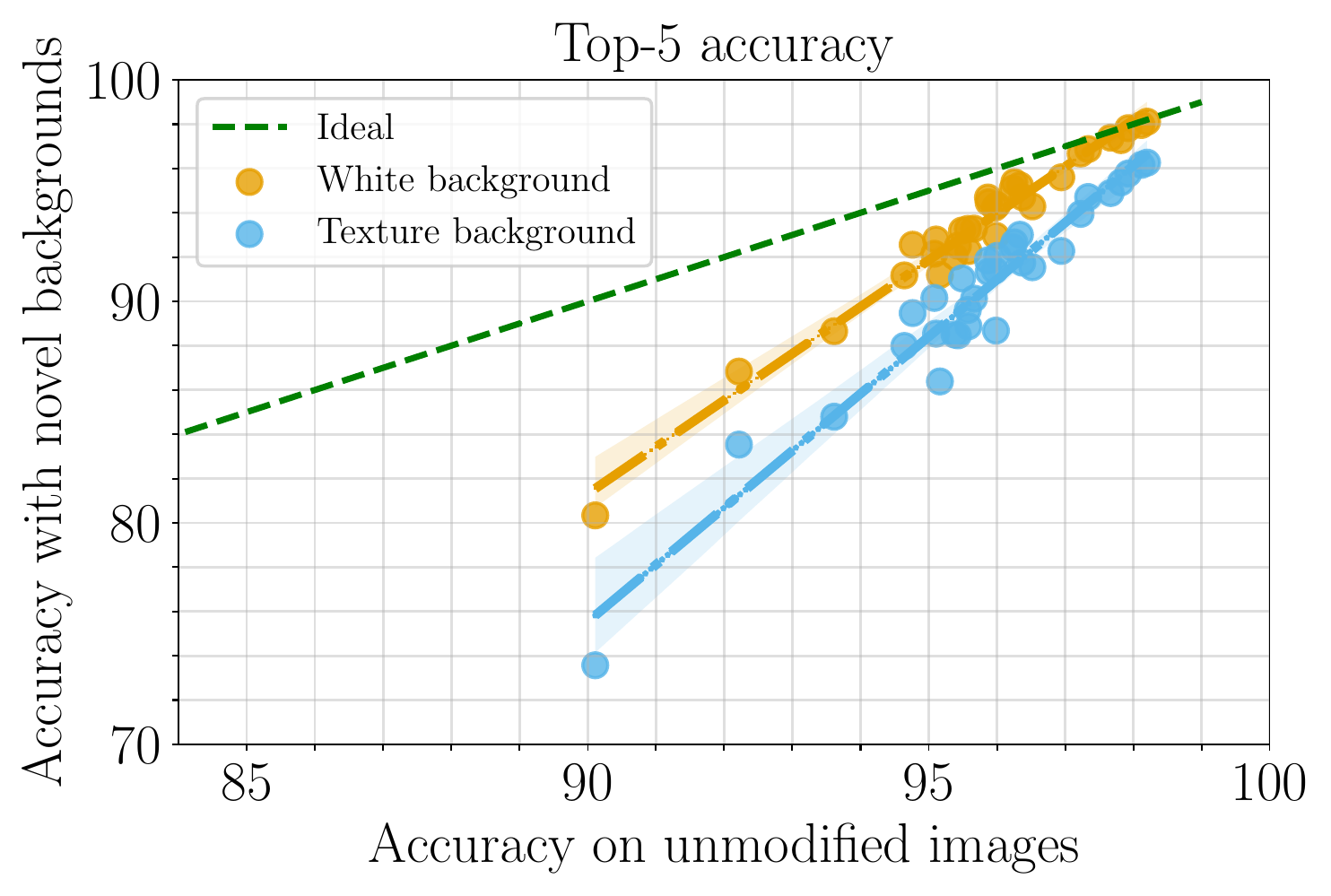}
    \caption{Top-5 accuracy on ImageNet test images when the background is switches to white or texture based background. Each dot correspond to one neural network.}
    \label{fig: switch_back_main_top5}
\end{figure}

Figure~\ref{fig: switch_back_main_top5} presents the top-5 accuracy for all 32 networks evaluated in this paper. Similar to the top-1 accuracy, our results show a large drop in performance when we switch background features. However, with the increasing expressive power of networks, this drop in performance decreases.

\end{document}